\pdfoutput=1

\documentclass[11pt]{article}

\usepackage[preprint]{acl}

\usepackage{times}
\usepackage{latexsym}

\usepackage[T1]{fontenc}

\usepackage[utf8]{inputenc}

\usepackage{microtype}

\usepackage{inconsolata}

\usepackage{graphicx}
\usepackage{algorithm}
\usepackage{algorithmic}
\usepackage{amsmath}

\title{Align-then-Slide: A complete evaluation framework for Ultra-Long Document-Level Machine Translation}

\author{
    Jiaxin GUO\thanks{These authors contributed equally to this work.}, 
    Daimeng Wei\footnotemark[1], Yuanchang Luo\footnotemark[1], 
    Xiaoyu Chen, Zhanglin Wu, Huan Yang, \\ 
    {\bf Hengchao Shang, Zongyao Li, Zhiqiang Rao, Jinlong Yang \and Hao Yang} \\
    Huawei Translation Services Center, Beijing, China \\
    \texttt{\{guojiaxin1,weidaimeng,luoyuanchang1\}@huawei.com}
}

\begin{document}
\maketitle
\begin{abstract}


Large language models (LLMs) have ushered in a new era for document-level machine translation (\textit{doc}-mt), yet their whole-document outputs challenge existing evaluation methods that assume sentence-by-sentence alignment. We introduce \textit{\textbf{Align-then-Slide}}, a complete evaluation framework for ultra-long doc-mt. In the Align stage, we automatically infer sentence-level source–target correspondences and rebuild the target to match the source sentence number, resolving omissions and many-to-one/one-to-many mappings. In the n-Chunk Sliding Evaluate stage, we calculate averaged metric scores under 1-, 2-, 3- and 4-chunk for multi-granularity assessment. Experiments on the WMT benchmark show a Pearson correlation of 0.929 between our method with expert MQM rankings. On a newly curated real-world test set, our method again aligns closely with human judgments. Furthermore, preference data produced by Align-then-Slide enables effective CPO training and its direct use as a reward model for GRPO, both yielding translations preferred over a vanilla SFT baseline. The results validate our framework as an accurate, robust, and actionable evaluation tool for doc-mt systems.

\end{abstract}

\section{Introduction}


Large language models (LLMs) are opening a new chapter for document-level machine translation ($doc$-mt) \cite{DBLP:conf/discomt/KimTN19,DBLP:journals/csur/MarufSH21,DBLP:conf/acl/FernandesYNM20}. Leveraging their exceptional capacity for long-context modeling and deep semantic understanding, LLMs can generate entire translations that are not only fluent and coherent but also faithful to the document’s global meaning, far surpassing the limitations of conventional sentence-by-sentence approaches.

In document-level machine translation evaluation, the prevailing paradigm is to lift proven sentence-level metrics such as BLEU \cite{DBLP:conf/acl/PapineniRWZ02}, BERTScore \cite{DBLP:journals/corr/abs-1904-09675} and COMET \cite{DBLP:conf/emnlp/ReiSFL20,DBLP:conf/wmt/ReiTGZFMSGACLM22} to the document level. The latest studies often use LLMs for scoring \cite{gu2025surveyllmasajudge}, but this method has biases, inaccuracies, and low efficiency \cite{guo2025automaticevaluationmetricsdocumentlevel}. \citeauthor{DBLP:conf/wmt/VernikosTMF22}’s $doc$-metrics achieves this with disarming simplicity: it merely prepends the preceding reference sentences to the current hypothesis–reference pair before encoding, instantly injecting document-wide context. \citeauthor{DBLP:conf/naacl/RaunakKP24}’s SLIDE adopts a chunk-wise strategy: a sliding window sweeps across the document, feeding contiguous blocks of sentences into an off-the-shelf quality-estimation model without any architectural tweaks, thereby enabling end-to-end document-level assessment.

\begin{figure}[t]
    \centering
    \includegraphics[width=0.9\columnwidth]{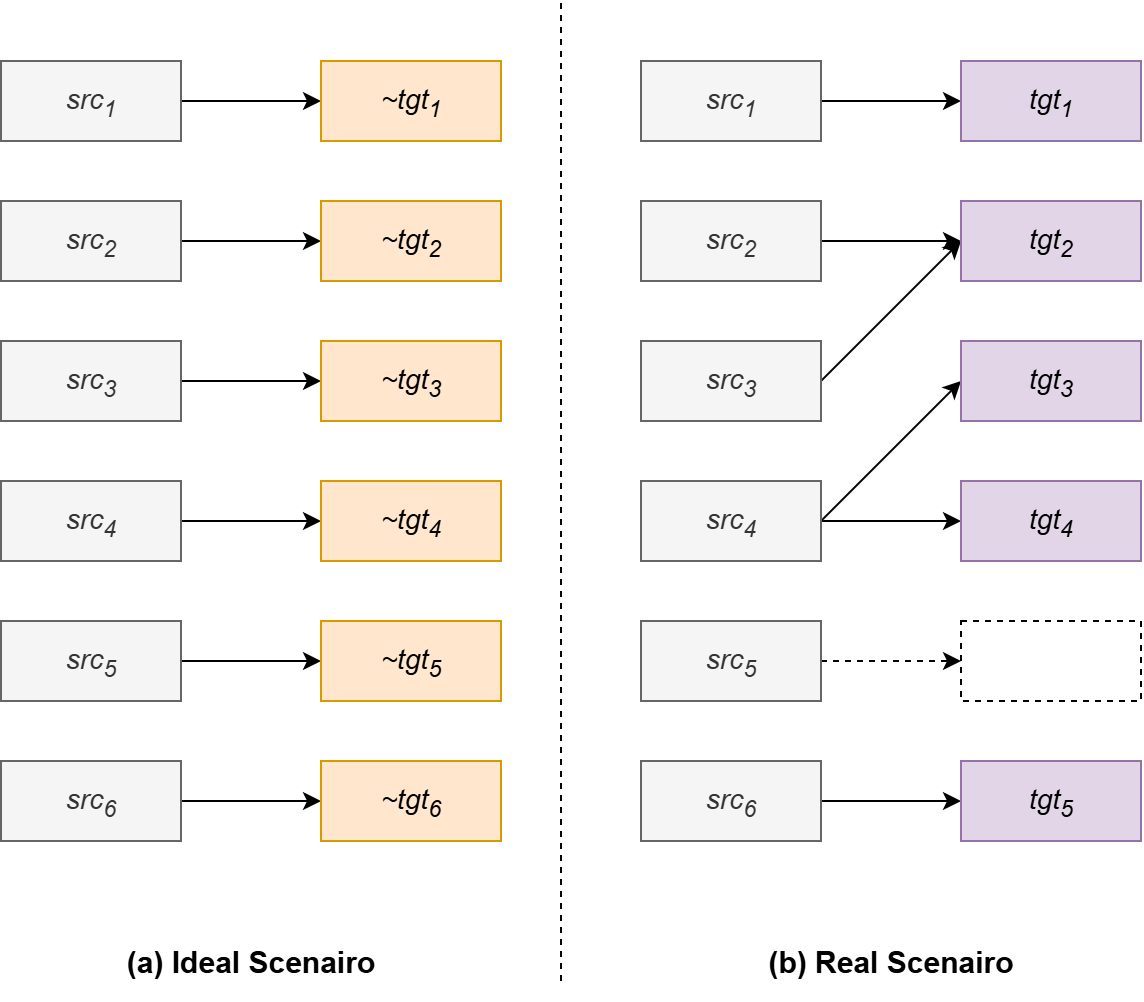}
    \caption{Ideal vs. Real scenarios between source and translated texts. (a) Perfect one-to-one correspondence assumed by prior metrics. (b) Actual complexities: whole-sentence omissions, many-to-one and one-to-many mappings, and variable target sentence counts.}
    \label{fig:case}
\end{figure}

\begin{figure*}[t]
    \centering
    \includegraphics[width=2.0\columnwidth]{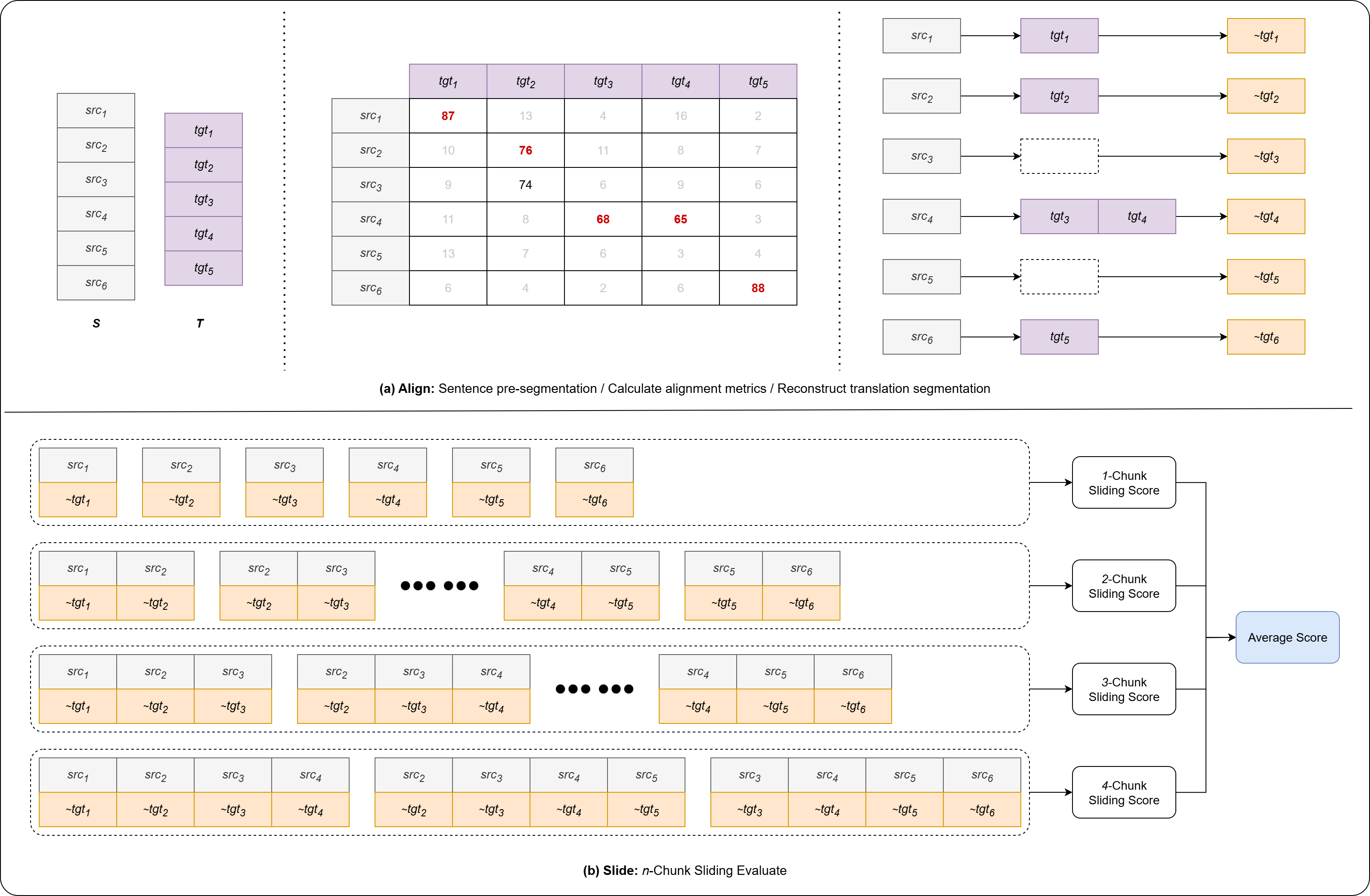}
    \caption{Overall pipeline of Align-then-Slide. (a) Align: constructing a one-to-one sentence-level correspondence between source and translation via optimal $dp$ search. (b) Slide: conducting n-Chunk sliding evaluation with stride one to assess quality at multiple granularities.}
    \label{fig:method}
\end{figure*}

Yet prior work implicitly assumes that the document has been segmented into sentences and that source and target sentences align one-to-one. Figure \ref{fig:case}(a) depicts this Ideal Scenario. The Real Scenario, shown in Figure \ref{fig:case}(b), introduces three thorny challenges:

\begin{enumerate}
    \item Whole sentence omission breaks the alignment. For example, $src_5$ is absent from the target.
    \item The mapping is no longer bijective: many-to-one ($src_2$ + $src_3$ → $tgt_2$) and one-to-many ($src_4$ → $tgt_3$ + $tgt_4$) relationships abound;
    \item Different systems generate varying numbers of target sentences, shattering the alignment assumption.
\end{enumerate}

In this paper we present \textit{\textbf{Align-then-Slide}}, a complete evaluation framework for ultra-long document-level machine translation. The approach unfolds in two stages.

\textit{\textbf{Stage 1}} \textbf{Align}: we first pre-segment both source and target texts and compute their sentence-level alignment matrix. Anchoring on the source sentence sequence, we then rebuild the target sequence so that the two have exactly the same number of sentences. During rebuilding, whole-sentence omissions are patched with placeholder sentences and one-to-many mappings are collapsed or expanded, all in one pass. This anchor-on-source design also neutralizes length discrepancies across different systems.

\textit{\textbf{Stage 2}} \textbf{$n$-Chunk Sliding Evaluate}: extending the spirit of BLEU’s $n$-grams to SLIDE\cite{DBLP:conf/naacl/RaunakKP24}, we compute metrics for 1-, 2-, 3-, and 4-chunk spans and average them. The 1-chunk scores sharply expose omissions, while 2- to 4-chunk scores mitigate the impact of many-to-one mappings, yielding a comprehensive, multi-granularity assessment.

\begin{algorithm*}[th]
\small
\caption{Stage 1: Align}
\label{alg:stage_1_align}
\begin{algorithmic}[1]
\REQUIRE {source document $S$, target document $T$, source language $src\_lang$, target language $tgt\_lang$}
\ENSURE {aligned source sentences $src\_lines$, reconstructed target sentences $new\_tgt\_lines$}

\STATE \textcolor{gray}{\texttt{// 1. Sentence pre-segmentation}}
\STATE $src\_lines \gets SEG(S, src\_lang)$
\STATE $tgt\_lines \gets SEG(T, tgt\_lang)$
\STATE $m \gets |{src\_lines}|$;\quad $n \gets |{tgt\_lines}|$

\STATE \textcolor{gray}{\texttt{// 2. Build $m\times n$ similarity matrix}}
\FOR{$i = 0 \ldots m-1$}
    \FOR{$j = 0 \ldots n-1$}
        \STATE ${score}[i][j] \gets \textit{EVAL}({src\_lines}[i], {tgt\_lines}[j])$ \hfill \COMMENT{e.g., COMET-\textit{Kiwi} or LaBSE}
    \ENDFOR
\ENDFOR

\STATE \textcolor{gray}{\texttt{// 3. Find optimal alignment path via DP}}
\STATE ${path} \gets \textit{DP\_SEARCH}({score})$ \hfill \COMMENT{Returns list of $(i,j)$ pairs}

\STATE \textcolor{gray}{\texttt{// 4. Reconstruct target aligned to source}}
\STATE ${new\_tgt\_lines} \gets [\ ]$
\FOR{$i = 0 \ldots m-1$}
    \STATE ${matched} \gets \{j \mid (i,j) \in {path}\}$
    \IF{${matched} = \emptyset$}
        \STATE ${new\_tgt\_lines}[i] \gets {""}$ \hfill \COMMENT{Placeholder for omission}
    \ELSE
        \STATE ${new\_tgt\_lines}[i] \gets \textit{CONCAT}({tgt\_lines}[j]\ \textbf{for}\ j\ \textbf{in}\ {matched})$
    \ENDIF
\ENDFOR

\STATE \RETURN ${src\_lines},\ {new\_tgt\_lines}$
\end{algorithmic}
\end{algorithm*}

We evaluate Align-then-Slide on the official WMT test suite, where its system-level ranking achieves a Pearson correlation of 0.929 with expert-based MQM\cite{Freitag_2021} scores, confirming its validity. Because this benchmark is already sentence-aligned and free of omissions, we further construct a realistic testbed by using various size LLMs to translate full documents and having professional translators rank the outputs. Align-then-Slide again shows strong agreement with human rankings. Moreover, the preference pairs generated by our metric can be fed directly into CPO \cite{xu2024contrastivepreferenceoptimizationpushing} training, or the metric itself can serve as a reward model for GRPO\cite{shao2024deepseekmathpushinglimitsmathematical}. Human evaluation reveals that both CPO- and GRPO-trained systems outperform a vanilla SFT (supervised fine-tuning) baseline, underscoring the reliability of our evaluation framework.

\section{Align-then-Slide}


In this paper, we present \textit{\textbf{Align-then-Slide}}, a comprehensive framework for evaluating ultra-long document-level machine translation. The method proceeds in two stages: Stage 1, \textit{\textbf{Align}}, establishes sentence-level correspondence between source and translation; Stage 2, \textit{\textbf{$n$-Chunked Sliding Evaluate}}, performs quality evaluation at multiple granularities.

\subsection{Align}
\label{sec:Align}
In this stage, we automatically segment the entire source and translated documents into a one-to-one set of aligned sentence pairs, as defined in Algorithm \ref{alg:stage_1_align}. The procedure is as follows:

\begin{itemize}
    \item Sentence pre-segmentation: independently segment both original and translated documents into sentence sequences.
    \item Calculate alignment metrics: compute sentence-level alignment similarity using reference-free metrics such as COMET-\textit{Kiwi}\cite{DBLP:conf/wmt/ReiSFL20} or LaBSE\cite{feng2022languageagnosticbertsentenceembedding}.
    \item Reconstruct translation segmentation: anchored on the source sentence order, we apply a dynamic-programming algorithm to merge or insert placeholder sentences, yielding a target sequence that exactly matches the source in length.
\end{itemize}

As shown in Figure \ref{fig:method}(a), for a source document $S$ and its translation $T$, we first segment both into sentences using off-the-shelf tools such as \textit{spaCy} and \textit{ersatz}, yielding $m$ source sentences $S = \{src_1, src_2, ..., src_m\}$ and $n$ target sentences $T = \{tgt_1, tgt_2, ..., tgt_n\}$. We then construct an $m × n$ similarity matrix, populated by reference-free metrics like COMET-\textit{Kiwi} or LaBSE. When $m = n$ and the mapping is one-to-one, the diagonal attains the maximum scores; in practice $m \ne n$, yet we can still find an optimal path that assigns each source sentence its best-matched target sentence. From the source perspective, unmatched positions are padded with placeholder sentences, while multiple matches are merged, resulting in a reconstructed target sequence $\{\tilde{tgt}_1, \tilde{tgt}_2, …, \tilde{tgt}_m\}$ of identical length $m$. Finding this optimal path is formulated as a dynamic programming problem as following:

\paragraph{Abstract problem description} An $[m, n]$ matrix with values at each point. Starting from $(0, 0)$ and ending at $(m-1, n-1)$, the path requirements are: the $y$-position must increase by \textit{1} each move, and the $x$-position must increase by a non-negative number each move, resulting in n points. The goal is to find a path that maximizes the sum of these $n$ points' values and provides their coordinates. We can use dynamic-programming (\textit{dp}) algorithm to achieve the goal. The algorithm steps are:
\begin{itemize}
    \item Initialize a 2-D array $dp$, where $dp[i][j]$ represents the maximum path value sum from $(0, 0)$ to $(m-1, n-1)$.
    \item Traverse the matrix. For each point $(i, j)$, calculate $dp[i][j]$. Given the $y$-position must increase by $1$ and the $x$-position by a non-negative number each move, $dp[i][j]$ is the maximum value from $dp[i-1][j-1], dp[i-1][j-2], ..., dp[i-1][0]$, plus the current point's value.
    \item Finally, $dp[m-1][n-1]$ is the maximum path value sum. Backtracking $dp$ gives the path's point coordinates.
\end{itemize}

As shown in Figure \ref{fig:method}(a), the optimal alignment path via \textit{dp} algorithm is highlighted in red: [($src_1$, $tgt_1$), ($src_2$, $tgt_2$), ($src_4$, $tgt_3$), ($src_4$, $tgt_4$), ($src_6$, $tgt_5$)]. Anchoring on the source sequence, we insert empty strings for missed sentences and concatenate multiple targets where necessary, resulting in the final mapping: [($src_1$, $tgt_1$), ($src_2$, $tgt_2$), ($src_3$, $""$), ($src_4$, $tgt_3$+$tgt_4$), ($src_5$, $""$), ($src_6$, $tgt_5$)]. The reconstructed target sequence $\{\tilde{tgt}_1, \tilde{tgt}_2, \tilde{tgt}_3, \tilde{tgt}_4, \tilde{tgt}_5, \tilde{tgt}_6\}$ is therefore $\{tgt_1, tgt_2, "", tgt_3+tgt_4, "", tgt_5\}$.

\begin{figure}[t]
    \centering
    \includegraphics[width=0.9\columnwidth]{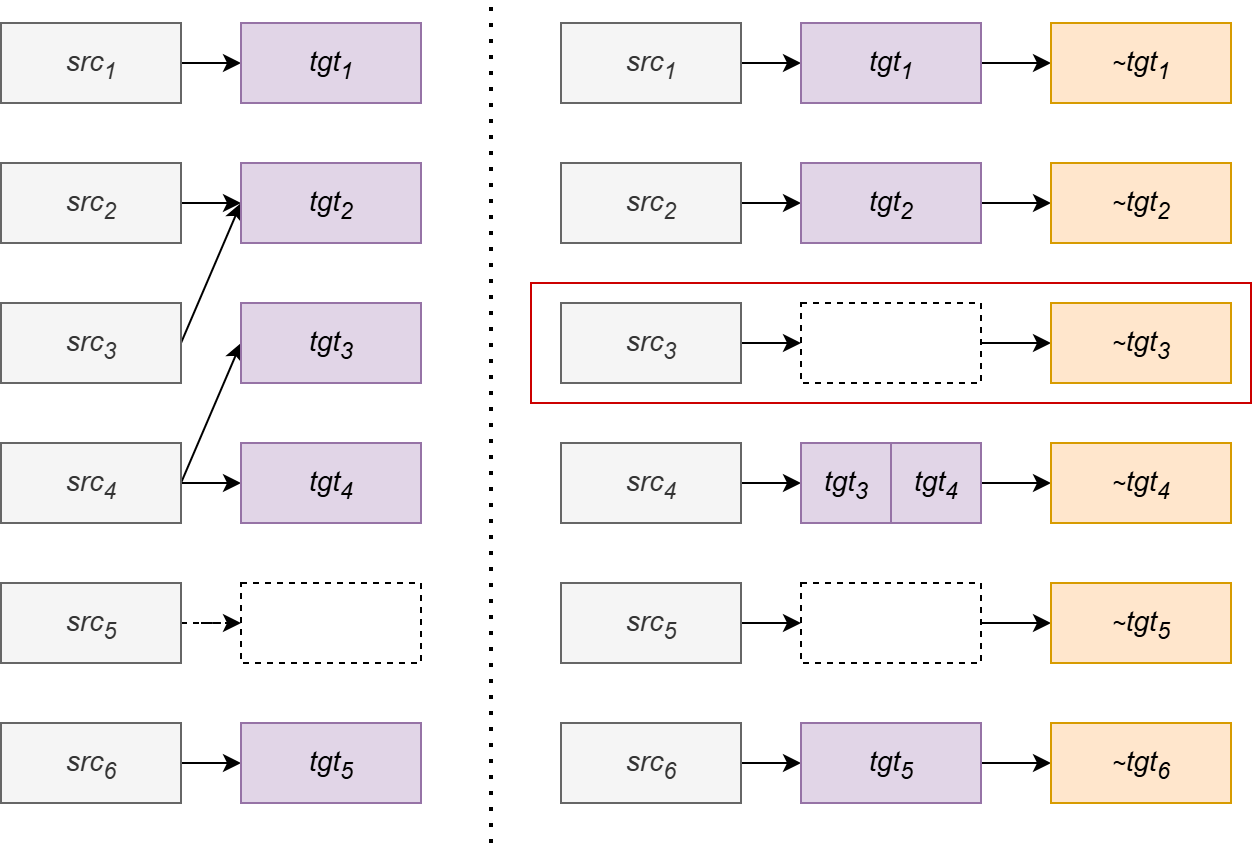}
    \caption{Alignment results after Stage 1. Sentence omissions (e.g., $src_5$) and one-to-many mappings (e.g., $src_4$) are resolved, whereas many-to-one conflicts (e.g., $src_3$ vs. $tgt_2$) remain and are addressed in Stage 2.}
    \label{fig:stage1_case}
\end{figure}

As shown in Figure \ref{fig:stage1_case}, Stage 1 \textit{Align} successfully resolves omissions (e.g., $src_5$) and one-to-many mappings (e.g., $src_4$). Yet many-to-one mappings introduce conflicts: $src_3$ should share $tgt_2$ but is instead marked as missing. Merging source sentences seems natural, yet it yields inconsistent segmentations across systems and undermines fair evaluation. We therefore defer this issue to Stage 2, where multi-granularity sliding windows neutralize the conflict.

\subsection{$n$-Chunk Sliding Evaluate}

Once sentence-level alignment is established, we conduct a multi-granularity sliding-window evaluation in this stage as shown in Figure \ref{fig:case}(b). 

Consecutive $k$ sentences form a chunk, and the window slides with a fixed stride of $1$, to heighten sensitivity to omissions. As shown in Figure \ref{fig:method}(b), for $m$ source sentences $S = \{src_1, src_2, ..., src_m\}$ and their aligned translations $T = \{\tilde{tgt}_1, \tilde{tgt}_2, ..., \tilde{tgt}_m\}$, we set $k \in \{1,2,3,4\}$, yielding $m-k+1$ units per setting; each unit is scored by a quality estimator, and the scores are averaged to produce the final metric score.

\begin{figure}[t]
    \centering
    \includegraphics[width=0.9\columnwidth]{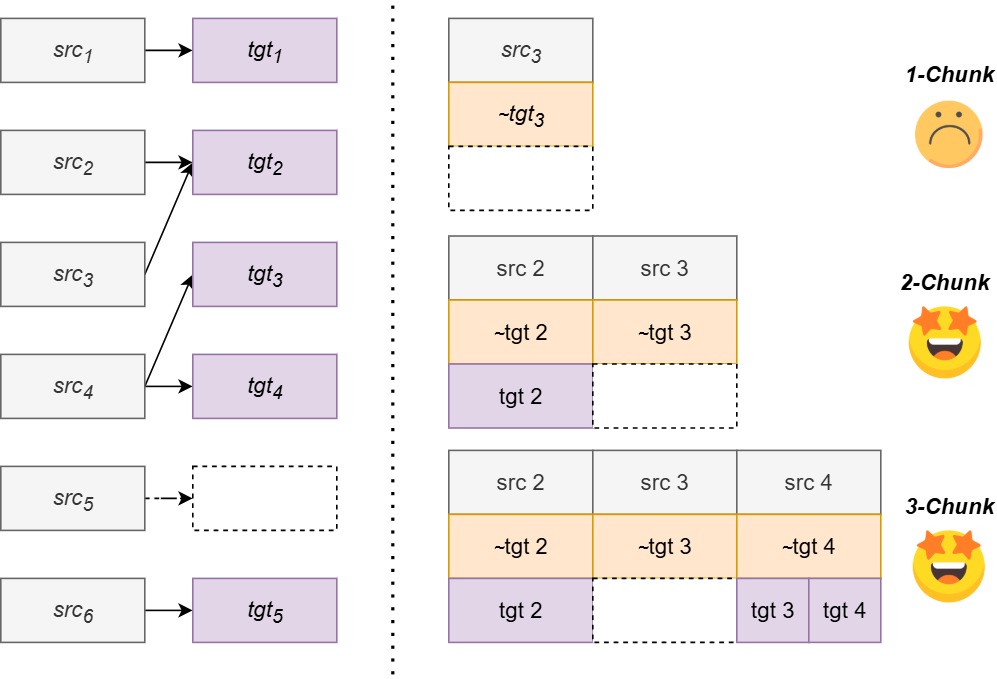}
    \caption{Multi-granularity n-chunk evaluation. While the 1-chunk unit flags $src_3$ as missing, 2-, 3-, and 4-chunk windows merge adjacent source sentences and restore correct alignment with $tgt_2$.}
    \label{fig:stage2_analysis}
\end{figure}

\paragraph{Why n-Chunk?}
Stage 1 \textit{Align} cannot handle many-to-one mappings: $src_3$ in Figure \ref{fig:stage1_case} is falsely judged empty because it shares $tgt_2$, causing 1-chunk scores to plummet. Merging source sentences directly would yield inconsistent segmentations across systems. We therefore shift the “merging” into the evaluation stage: when chunk>1, adjacent source sentences are grouped into a single unit, allowing the translation to be re-matched at coarser granularities. As shown in Figure \ref{fig:stage2_analysis}, $src_3$ is correctly aligned within 2-, 3-, and 4-chunk units. 

The key distinction from SLIDE lies in our introduction of a hierarchical chunk-based evaluation strategy coupled with a fixed sliding stride of $1$, whereas SLIDE employs a dynamic sliding window.


\section{Experiments}

We conduct two sets of experiments to validate \textit{Align-then-Slide}.
\begin{enumerate}
    \item \textbf{Correlation study:} we measure the agreement between our metric and human judgments on both the standard WMT test set and our newly curated real-world benchmark.
    \item \textbf{Training efficacy study:} we fairly compare the quality of translations produced by vanilla supervised fine-tuning (SFT) versus those refined via reinforcement learning (CPO / GRPO) guided by our evaluation strategy.
\end{enumerate}

\subsection{Correlation Study Setup}
\subsubsection{TestSets}
\paragraph{Standard Testsets} We adopt the WMT 2020 Chinese$\rightarrow$English tracks as our standard benchmarks. Each track contains submissions from multiple participating teams. We re-assemble the sentence-level outputs into full documents and evaluate them with \textit{Align-then-Slide}, yielding a system ranking., yielding a system ranking. The resulting system rankings are compared with expert-based MQM rankings via Pearson correlation.

\paragraph{Real-world Testsets} Standard test sets are sentence-aligned and therefore do not reflect authentic document-level translations. We therefore constructed new Chinese$\rightarrow$English and English$\rightarrow$Chinese test sets, each containing outputs from six Qwen\footnote{https://huggingface.co/Qwen} LLMs; construction details are provided in Appendix \ref{app:real_world_testsets}. Professional translators produced pairwise relative rankings of these model outputs, establishing human system ranks. We then applied Align-then-Slide to the same outputs to obtain automatic ranks and report the Pearson correlation between the two.

\subsubsection{Align-then-Slide Setup}
\label{sec:sec_setup}

For Stage 1, sentence pre-segmentation is performed with \textit{spaCy}\footnote{https://spacy.io/models}, and alignment scores are computed using COMET\textit{Kiwi}. In Stage 2, we instantiate three variants by plugging different COMET backbones, namely COMET20\footnote{https://huggingface.co/Unbabel/wmt20-comet-da}, COMET22\footnote{https://huggingface.co/Unbabel/wmt22-comet-da}, and COMET\textit{Kiwi}\footnote{https://huggingface.co/Unbabel/wmt22-cometkiwi-da}, denoted ASD20, ASD22, and ASD\textit{Kiwi}, respectively. We also experiment with \textit{ersatz}\cite{wicks-post-2021-unified} for pre-segmentation and LaBSE for alignment; these ablations are reported in Section \ref{sec:ablation}.

\subsection{Training Efficacy Study Setup}

Post-training of LLMs typically follows two paradigms: supervised fine-tuning (SFT) and reinforcement learning (RL). For \textit{doc}-mt, the literature has almost exclusively adopted SFT, as parallel document pairs can be readily distilled from state-of-the-art models. While a handful of studies have explored RL, they remain confined to sentence-level tasks, where quality estimation is mature. Owing to the absence of reliable document-level metrics, RL training for full-document translation has been largely unexplored.

We benchmark SFT against RL on the Qwen2.5-7B backbone for both Chinese$\rightarrow$English (ZH$\rightarrow$EN) and English$\rightarrow$Chinese (EN$\rightarrow$ZH)to verify the training utility of \textit{Align-then-Slide}.

\paragraph{Data} We collect 50k document-level bilingual pairs. Given SFT’s sensitivity to quality, we first distill the raw corpus with Qwen3-32B and Qwen2.5-72B, yielding $D_{3-32}$ and $D_{2.5-72}$. We then select the higher-scoring translation for each source via \textit{Align-then-Slide} to obtain $D_{best}$.

\paragraph{SFT Training} We train $M_{sft-3-32}$, $M_{sft-2.5-72}$ and $M_{sft-best}$ on $D_{3-32}$, $D_{2.5-72}$ and $D_{best}$.

\paragraph{RL Training}
\begin{itemize}
    \item Offline RL (CPO) creates preference triplets by scoring the two distilled outputs with \textit{Align-then-Slide}, yielding $D_{cpo}$ and model $M_{cpo}$.
    \item Online RL (GRPO) trains on $D_{best}$ with \textit{Align-then-Slide} as the reward model, sampling eight translations per step to produce $M_{grpo}$.
\end{itemize}

\paragraph{Evaluation} All five models are compared on the same test set.


\section{Results and Analysis}

\subsection{Correlation Study Results}

\subsubsection{Results for Standard Testsets}

Table \ref{tab:rst1_p} summarizes the alignment between ASD20 rankings and official MQM rankings for seven systems on the WMT2020 Chinese$\rightarrow$English test set, quantified by both \textit{Pearson} and \textit{Kendall}. Table \ref{tab:rst1_detail} presents the detailed scores. For comparison, we also list sentence-level COMET20 scores and rankings. All MQM scores and ranks for the seven systems are taken from the official WMT release\footnote{https://github.com/google/wmt-mqm-human-evaluation}.

\begin{table}[h]
  \centering
  \begin{tabular}{lcc}
    \hline
    \textbf{} & \textbf{Pearson} & \textbf{Kendall} \\
    \hline
    MQM & 1 & 1 \\
    COMET20* & 0.679 & 0.524 \\
    ASD20 & 0.929 & 0.810 \\
    \hline
  \end{tabular}
  \caption{Correlation between ASD20 rankings and official MQM rankings on the WMT2020 Chinese$\rightarrow$English test set for seven systems, measured by Pearson and Kendall.}
  \label{tab:rst1_p}
\end{table}

\begin{table}[h]
  \centering
  \begin{tabular}{lccc}
    \hline
    \textbf{System} & \textbf{MQM} & \textbf{COMET20} & \textbf{ASD20} \\
    \hline
    VolcTrans            & 5.03(1) & 0.509(3) & 0.490(2) \\
    Wechat\_AI           & 5.13(2) & 0.522(1) & 0.496(1) \\
    Tencent              & 5.19(3) & 0.511(2) & 0.487(3) \\
    OPPO                 & 5.20(4) & 0.500(5) & 0.482(5) \\
    THUMT                & 5.34(5) & 0.497(6) & 0.485(4) \\
    DeepMind             & 5.41(6) & 0.493(7) & 0.480(6) \\
    DiDiNLP              & 5.48(7) & 0.502(4) & 0.477(7) \\
    \hline
  \end{tabular}
  \caption{Detailed scores and rankings for seven systems on the WMT2020 Chinese$\rightarrow$English test set, including ASD20, sentence-level COMET20, and official MQM values.}
  \label{tab:rst1_detail}
\end{table}

Table \ref{tab:rst1_p} shows that ASD20 correlates strongly with MQM, achieving a Pearson coefficient of 0.929 and a Kendall’s of 0.81. Table \ref{tab:rst1_detail} reveals that the two rankings differ only at minor positions, VolcTrans and Wechat\_AI swap the 1st and 2nd spots, while OPPO and THUMT exchange the 4th and 5th. In contrast, sentence-level COMET20 rankings deviate substantially from the MQM order.

We substituted the evaluation with ASD22 and ASDKiwi and repeated the experiments, again obtaining similar and consistent outcomes, see Appendix \ref{app:rst_standard}. These findings demonstrate the broad applicability of the \textit{Align-then-Slide} framework.

\subsubsection{Results for Real-world Testsets}

\begin{table}[h]
  \centering
  \begin{tabular}{lcc}
    \hline
    \textbf{System} & \textbf{Rank} & \textbf{ASD20} \\
    \hline
    Qwen3-32B    & 1 & 0.5203(1) \\
    Qwen2.5-72B  & 2 & 0.5181(2) \\
    Qwen3-8B     & 3 & 0.5041(4) \\
    Qwen2.5-32B  & 4 & 0.5096(3) \\
    Qwen2.5-14B  & 5 & 0.4939(5) \\
    Qwen2.5-7B   & 6 & 0.4906(6) \\
    \hline
  \end{tabular}
  \caption{Agreement between ASD20 rankings and human rankings for six LLMs on the Real-world Chinese$\rightarrow$English Testsets.}
  \label{tab:rst2_rank}
\end{table}

Table~\ref{tab:rst2_rank} compares the rankings produced by \textit{Align-then-Slide} with human judgments across six LLMs on our Real-World Chinese$\rightarrow$English Testsets. Remarkably, ASD20 aligns almost perfectly with the human order, misplacing only two systems, and attains a Pearson correlation of 0.943. This provides strong additional evidence for the validity of \textit{Align-then-Slide} in \textit{doc}-mt evaluation.

We repeated the experiment on the Real-world Testsets with ASD22 and ASD\textit{Kiwi}, and further extended it to English$\rightarrow$Chinese. All runs achieved similarly high agreement with human rankings. Details are in Appendix \ref{app:rst_real}. These results underscore the universality of \textit{Align-then-Slide}.

\subsection{Training Efficacy Study Results}

Table \ref{tab:rst3_train_rank} reports human-ranked system quality: $M_{grpo}$ $\approx$ $M_{cpo}$ > $M_{sft-best}$ > $M_{sft-2.5-72}$ $\approx$ $M_{sft-3-32}$.

\begin{table}[h]
  \centering
  \begin{tabular}{lcc}
    \hline
    \textbf{Model} & \textbf{ZH$\rightarrow$EN} & \textbf{EN$\rightarrow$ZH} \\
    \hline
    Baseline          & 6 & 6 \\
    $M_{sft-3-32}$    & 5 & 4 \\
    $M_{sft-2.5-72}$  & 4 & 5 \\
    $M_{sft-best}$    & 3 & 3 \\
    $M_{cpo}$         & 1 & 2 \\
    $M_{grpo}$        & 2 & 1 \\
    \hline
  \end{tabular}
  \caption{Human evaluation rankings on the Chinese→English and English→Chinese test sets.}
  \label{tab:rst3_train_rank}
\end{table}

This demonstrates two key points. First, RL training guided by \textit{Align-then-Slide} (both GRPO and CPO) significantly outperforms all SFT baselines, confirming the framework’s effectiveness for document-level RL. Second, among the SFT models, the one trained on $D_{best}$, selected via Align-then-Slide—achieves the highest quality, showing that $D_{best}$ is consistently superior to the single-model distillations $D_{sft-2.5-72}$ and $D_{sft-3-32}$. Thus, Align-then-Slide not only steers training but also reliably identifies higher-quality data, underscoring its utility across the entire training pipeline.

\section{Ablation Study}
\label{sec:ablation}

\begin{table*}
  \centering
  \begin{tabular}{lcccc}
    \hline
    \textbf{System} & \textbf{Rank} & \textbf{spaCy} + \textbf{COMET\textit{Kiwi}} & \textbf{ersatz} + \textbf{COMET\textit{Kiwi}}  & \textbf{spaCy} + \textbf{Labse}  \\
    \hline
    Qwen3-32B    & 1 & 0.5203(1)  & 0.5201(1) & 0.5203(1)   \\
    Qwen2.5-72B  & 2 & 0.5181(2)  & 0.5180(2) & 0.5181(2)   \\
    Qwen3-8B     & 3 & 0.5041(4)  & 0.5041(4) & 0.5041(4)   \\
    Qwen2.5-32B  & 4 & 0.5096(3)  & 0.5095(3) & 0.5096(3)   \\
    Qwen2.5-14B  & 5 & 0.4939(5)  & 0.4936(6) & 0.4940(5)   \\
    Qwen2.5-7B   & 6 & 0.4906(6)  & 0.4911(5) & 0.4906(6)   \\
    \hline
  \end{tabular}
  \caption{Ablation results on segmentation tools (\textit{spaCy} vs.\ \textit{ersatz}) and similarity metrics (COMET\textit{Kiwi} vs.\ LaBSE) within the alignment stage of Align-then-Slide.}
  \label{tab:ablation}
\end{table*}

Table \ref{tab:ablation} ablates the two pivotal components of the alignment stage: the document pre-segmentation tool and the model that produces the m×n similarity matrix. We compare \textit{spaCy} and \textit{ersatz} for segmentation, and COMET\textit{Kiwi} and LaBSE for alignment modeling, all introduced in Section \ref{sec:sec_setup}. 

The three columns of Table \ref{tab:ablation} correspond to these configurations: the original “\textit{spaCy} + COMET\textit{Kiwi}”, the segmentation variant “\textit{ersatz} + COMET\textit{Kiwi}”, and the similarity-metric variant “\textit{spaCy} + LaBSE”.

\paragraph{Segment Tools}

Comparing the “\textit{spaCy} + COMET\textit{Kiwi}” and “\textit{ersatz} + COMET\textit{Kiwi}” columns reveals that ASD20 scores for each system differ only marginally, and the predicted rankings remain identical. This stability stems from the maturity of current segmentation tools: a spot-check of 120 sentences shows just seven (5.83\%) segmentation mismatches. Because both hypotheses and references are aligned to the source segmentation, these minor tool differences have negligible impact on the final evaluation.

\paragraph{Alignment Models}

After segmentation, alignment hinges solely on the $m \times n$ similarity matrix. Comparing “\textit{spaCy} + COMET\textit{Kiwi}” and “\textit{spaCy} + LaBSE” shows near-identical ASD20 scores and identical rankings across systems. This stems from both the high accuracy of current alignment models and our DP algorithm’s ability to find a consistently optimal path, underscoring the robustness of the ASD approach.

\section{Conclusion}

Targeting \textit{doc}-mt evaluation challenges, this paper proposes an align-then-slide evaluation method, forming a complete metric system. By automatically constructing sentence-level alignment and combining it with $n$-chunk sliding evaluation, it overcomes traditional metric limitations and offers a full solution for \textit{doc}-mt evaluation. Future work will focus on further algorithm optimization to enhance the metrics' accuracy and efficiency.

\section{Limitations}

Computational Cost. Generating the $m \times n$ similarity matrix and performing sliding-window evaluation incurs $\mathcal{O}(m \times n)$ memory and $\mathcal{O}(k \times (m - k + 1))$ extra calls to the backbone model. For very long documents, GPU memory and latency can become prohibitive without batching or pruning heuristics.

\bibliography{custom}

\clearpage
\onecolumn
\appendix

\section{Appendix 1: Real-world Testsets}
\label{app:real_world_testsets}

Our bilingual data originate from CommonCrawl. We first randomly sampled 100 document pairs that contained both source and target texts. After rule-based filtering to remove poorly aligned samples, professional translators selected the 50 highest-quality pairs to form our real-world test set.

\section{Appendix 2: Results for Standard Testsets}
\label{app:rst_standard}

\begin{table}[h]
  \centering
  \begin{tabular}{lcc}
    \hline
    \textbf{} & \textbf{Pearson} & \textbf{Kendall} \\
    \hline
    ASD20 & 0.929 & 0.810 \\
    ASD22 & 0.893 & 0.714 \\
    ASD\textit{Kiwi} & 0.964 & 0.905 \\
    \hline
  \end{tabular}
  \caption{Correlation between ASD20 / ASD22/ ASD\textit{Kiwi} rankings and official MQM rankings on the WMT2020 test set for seven systems, measured by Pearson and Kendall.}
  \label{tab:app_rst_standard}
\end{table}

\section{Appendix 3: Results for Real-world Testsets}
\label{app:rst_real}

\begin{table}[h]
  \centering
  \begin{tabular}{lcccc}
    \hline
    \textbf{System} & \textbf{Rank} & \textbf{ASD20} & \textbf{ASD22} & \textbf{ASD\textit{Kiwi}} \\
    \hline
    Qwen3-32B    & 1 & 0.5203(1) & 0.8378(2) & 0.8204(2) \\
    Qwen2.5-72B  & 2 & 0.5181(2) & 0.8385(1) & 0.8207(1) \\
    Qwen3-8B     & 3 & 0.5041(4) & 0.8362(3) & 0.8190(4) \\
    Qwen2.5-32B  & 4 & 0.5096(3) & 0.8345(4) & 0.8192(3) \\
    Qwen2.5-14B  & 5 & 0.4939(5) & 0.8304(5) & 0.8187(5) \\
    Qwen2.5-7B   & 6 & 0.4906(6) & 0.8293(6) & 0.8184(6) \\
    \hline
  \end{tabular}
  \caption{Agreement between ASD20 / ASD22/ ASD\textit{Kiwi} rankings and human rankings for six LLMs on the Real-world Chinese$\rightarrow$English Testsets.}
  \label{tab:app_rst_real1}
\end{table}

\begin{table}[h]
  \centering
  \begin{tabular}{lcccc}
    \hline
    \textbf{System} & \textbf{Rank} & \textbf{ASD20} & \textbf{ASD22} & \textbf{ASD\textit{Kiwi}} \\
    \hline
    Qwen3-32B    & 1 & 0.3712(2) & 0.7261(1) & 0.8245(2) \\
    Qwen2.5-72B  & 2 & 0.3746(1) & 0.7255(2) & 0.8287(1) \\
    Qwen2.5-32B  & 3 & 0.3502(3) & 0.7125(4) & 0.8189(4) \\
    Qwen3-8B     & 4 & 0.3486(4) & 0.7134(3) & 0.8203(3) \\
    Qwen2.5-14B  & 5 & 0.3361(5) & 0.7018(5) & 0.8138(5) \\
    Qwen2.5-7B   & 6 & 0.3144(6) & 0.6910(6) & 0.7325(6) \\
    \hline
  \end{tabular}
  \caption{Agreement between ASD20 / ASD22/ ASD\textit{Kiwi} rankings and human rankings for six LLMs on the Real-world English$\rightarrow$Chinese Testsets.}
  \label{tab:app_rst_real2}
\end{table}

\end{document}